\documentclass{article}

\PassOptionsToPackage{numbers, compress}{natbib}

    \usepackage[preprint]{neurips_2025}

\usepackage[utf8]{inputenc} 
\usepackage[T1]{fontenc}    
\usepackage{hyperref}       
\usepackage{url}            
\usepackage{booktabs}       
\usepackage{amsfonts}       
\usepackage{nicefrac}       
\usepackage{microtype}      
\usepackage{xcolor}         
\usepackage{pifont}
\usepackage{graphicx}
\usepackage{multirow}
\usepackage{multicol}
\usepackage{enumitem}
\usepackage{wrapfig}
\usepackage{listings}
\newcommand{\cmark}{{\color{green!70!black}\ding{51}}} 
\newcommand{\xmark}{{\color{red}\ding{55}}}            

\title{GUI-Robust: A Comprehensive Dataset for Testing GUI Agent Robustness in Real-World Anomalies}

\newcommand{\ie}{\emph{i.e., }}
\newcommand{\eg}{\emph{e.g., }}

\author{%
  \And
  Jingqi Yang\thanks{Contact: \texttt{chessbean1@gmail.com}}  \\
  Zhejiang University \\
  \And
  Zhilong Song \\
  Intelligence Indeed \\
  \And
  Jiawei Chen \\
  Zhejiang University \\
  \And
  Mingli Song \\
  Zhejiang University \\
  \And
  Sheng Zhou \\
  Zhejiang University \\
  \And
  linjun sun \\
  Intelligence Indeed \\
  \And
  Xiaogang Ouyang \\
  Intelligence Indeed \\
  \And
  Chun Chen \\
  Zhejiang University \\
  \And
  Can Wang \\
  Zhejiang University \\
}

\begin{document}

\maketitle

\begin{abstract}

  The development of high-quality datasets is crucial for benchmarking and advancing research in Graphical User Interface (GUI) agents.  Despite their importance, existing datasets are often constructed under idealized conditions, overlooking the diverse anomalies frequently encountered in real-world deployments. To address this limitation, we introduce GUI-Robust, a novel dataset designed for comprehensive GUI agent evaluation, explicitly incorporating seven common types of anomalies observed in everyday GUI interactions. Furthermore, we propose a semi-automated dataset construction paradigm that collects user action sequences from natural interactions via RPA tools and then generate corresponding step and task descriptions for these actions with the assistance of MLLMs. This paradigm significantly reduces annotation time cost by a factor of over 19 times. Finally, we assess state-of-the-art GUI agents using the GUI-Robust dataset, revealing their substantial performance degradation in abnormal scenarios. We anticipate that our work will highlight the importance of robustness in GUI agents and inspires more future research in this direction. The dataset and code are available at \url{https://github.com/chessbean1/GUI-Robust}. 
  
\end{abstract}

\section{Introduction}

A GUI agent is an intelligent system capable of autonomously interacting with graphical user interfaces by perceiving visual elements, understanding task objectives, and executing corresponding actions ~\cite{nguyen2024guiagentssurvey}. The development of GUI agents holds significant promise for automating GUI operations, reducing human workload and enabling seamless human-computer collaboration. Traditionally, GUI automation has relied on heuristic rules, brittle scripts, or hard-coded templates, which limit flexibility and generalization ~\cite{rule-based}~\cite{tool-based}. Recently, the emergence of Multimodal Large Language Models (MLLMs) ~\cite{Qwen2.5-VL}~\cite{openai2024gpt4ocard}~\cite{geminiteam2024geminifamilyhighlycapable} has accelerated GUI agent research.  By integrating visual perception with language understanding, MLLMs enable agents to interpret complex screen layouts, comprehend natural language instructions, and reason about sequential actions in open-ended environments ~\cite{chin2024humancenteredllmagentuserinterface} --- marking a significant advance toward fully GUI automation.

In this rapidly evolving field, the construction of high-quality datasets is essential for benchmarking and advancing GUI agent research. However, existing datasets are typically constructed under idealized conditions and fail to capture the diverse range of anomalies encountered in real-world deployment. In practice, GUI agents deployed in industrial or consumer applications frequently encounter unexpected failure modes and environmental disturbances, such as action failures, obstructive pop-up advertisements, network disconnections, etc.  These anomalies can significantly disrupt execution flow, leading to task failures or, in severe cases, unintended interactions with sensitive interface components, potentially resulting in erroneous or hazardous outcomes. 

\textbf{New Dataset: GUI-Robust.} To address this gap, we present \textbf{GUI-Robust}, the first dataset designed to evaluate robustness in the presence of abnormal scenarios. GUI-Robust contains {5,318 annotated tasks} (consists of task descriptions and user behaviors sequences) collected from {392 diverse sources}, spanning both websites and third-party desktop applications on Windows. Notably, it includes 200 abnormal tasks covering 7 common types of anomalies encountered in everyday GUI usage including action failure, login page, captcha page, ad pop-up, cookie pop-up, page loading and network disconnection. This enables rigorous evaluation of GUI agent robustness against real-world anomalies, a critical aspect for practical deployment.

Beyond its unique property of robustness, as shown in Table ~\ref{tab:benchmark-comparison}, GUI-Robust offers several advantages for comprehensive evaluation: (1) it includes a broad range of task and action types including click, input text, retrieve information from page, open a new web or app and report anomalies to human; (2) it incorporates cross-scenario tasks spanning multiple applications or websites, reflecting more realistic and complex workflows; and (3) it covers both Chinese and English software environments. A comparison between GUI-Robust and existing benchmarks is shown in Table~\ref{tab:benchmark-comparison}. These features ensure that our dataset closely aligns with real-world usage, supporting thorough assessment of agent robustness, adaptability, and generalization.

\textbf{Novel Data Collection Strategy: RevAct.} Traditional dataset construction typically relies on a manual pipeline --- ranging from task design to execution planning and manual demonstration --- which is labor-intensive, costly, and requires significant domain expertise. To address these limitations, we propose a novel data collection strategy, {RevAct}, which reverses the conventional workflow and enables semi-automated dataset generation: we first collect user action sequences from natural interactions via RPA(Robotic Process Automation) tools, and then generate specific step and task descriptions for these actions with the assistance of MLLMs. This approach substantially reduces annotation costs, as expert involvement is limited to reviewing and revising step and task descriptions generated by MLLMs. Specifically, for each user action sequence, we leverage YOLOv8 for GUI element detection and Qwen2.5-VL for task generation and summarization, achieving over 71\% accuracy in automatic task generation. Human annotators are only required for minimal correction, reducing annotation time by a factor of over 19 times compared to the traditional pipeline.

\begin{figure}[t]  
    \centering  
    \includegraphics[width=\textwidth]{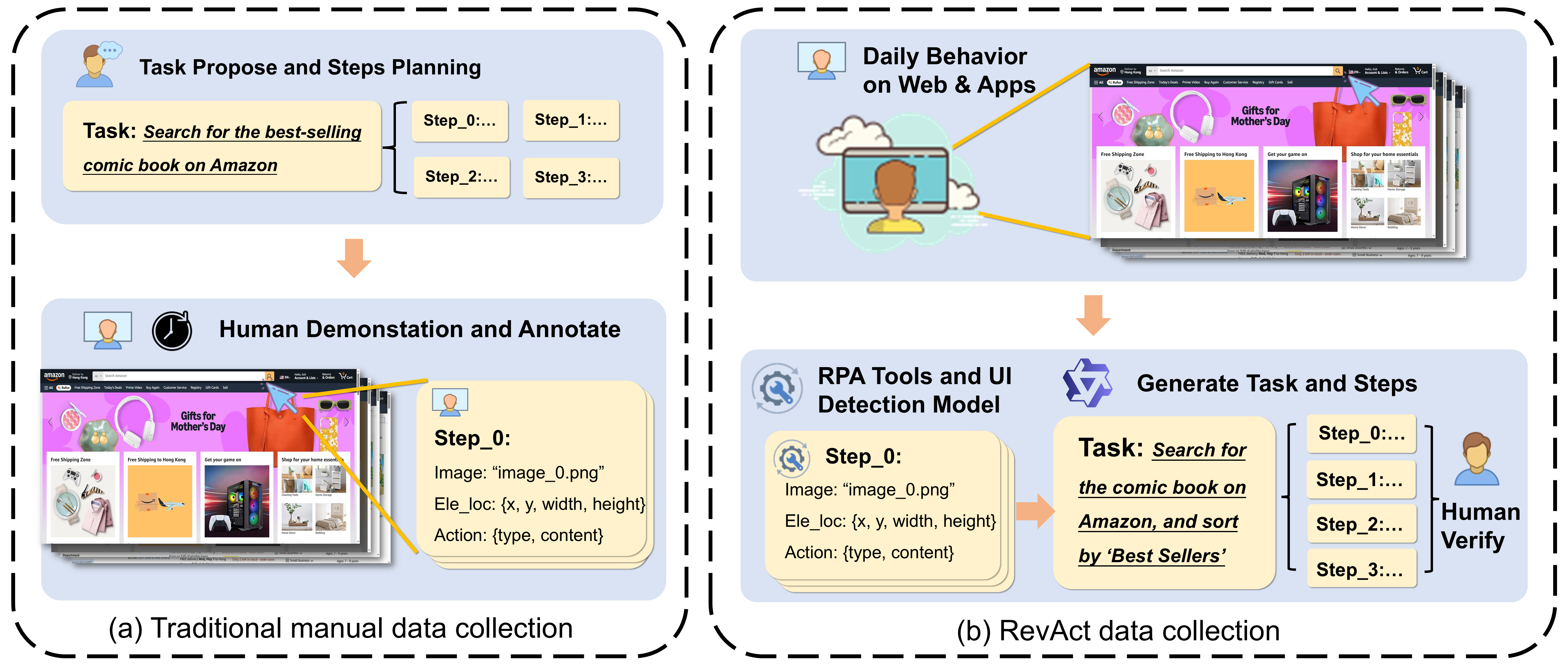}
    \caption{Comparison between  (a) traditional manual data collection and (b) our proposed RevAct pipeline. Manual pipelines require annotators to explicitly design tasks, plan execution steps, demonstrate actions, and annotate each step. In contrast, RevAct records natural user behavior on web/apps, leverages automated tools (e.g., RPA, detection models, and LLMs) to infer action semantics and generate step/task descriptions, at last verify by human.}  
    \label{fig:comparison}  
     \vspace{-0.4cm}
\end{figure}

\textbf{Comprehensive Experiments.} GUI-Robust provides a thorough benchmark for evaluating GUI agents on element grounding, multi-step task completion, cross-scenario execution, and robustness under abnormal conditions. We conduct extensive experiments with three representative MLLMs and two specific-designed GUI agents, demonstrating that all models experience significant performance degradation in abnormal scenarios. This underscores that robust GUI understanding in real-world settings remains a challenging problem. We hope our work will draw greater attention to the robustness of GUI agents and inspire further research in this direction based on our dataset.

\section{Ralated Work}
\begin{table}[t]
  \centering
  \caption{Statistics of GUI-Robust compared with existing datasets.}
  \label{tab:benchmark-comparison}
  \resizebox{\textwidth}{!}{
  \begin{tabular}{lcccccccc}
    \toprule
    \textbf{Benchmark} & \textbf{\#Apps/Web} & \textbf{\#Tasks} & \textbf{Platform} & \textbf{Abnormal?} & \textbf{Cross?} & \textbf{Lang.}   \\
    \midrule
    Mind2Web  & 137     & 2,350   & Desktop Web       & \xmark & \xmark & EN        \\
    WebLINX   & 155     & 2,337   & Desktop Web       & \xmark & \xmark & EN        \\
    OSWorld  & 9     & 369   & Desktop(Apps+Web) & \xmark & \cmark & EN        \\
    WindowsAgentArena  & 11     & 154   & Desktop(Apps+Web) & \xmark & \xmark & EN        \\
    GUI Odyssey  & 201     & 7,735   & Android Apps & \xmark & \cmark & EN       \\
    SPA-Bench & 58      & 340    & Android Apps      & \xmark & \cmark & EN \& CN   \\
    WorldGUI  & 10      & 315    & Desktop Apps      & \xmark & \xmark & EN       \\
    \midrule
    GUI-Robust & 392    & 5,318  & Desktop(Apps+Web) & \cmark & \cmark & EN \& CN  \\
    \bottomrule
  \end{tabular}
  }
\end{table}

\subsection{GUI Agents}

Early progress in GUI agent research has been largely driven by the rapid advancement of multimodal large language models (MLLMs), such as GPT-4~\cite{openai2024gpt4technicalreport}, Gemini-2.5~\cite{geminiteam2024geminifamilyhighlycapable}, Qwen-VL~\cite{Qwen-VL} and Claude3.5. These foundation models—capable of processing both visual and textual inputs—enable agents to interpret screen layouts, follow natural language instructions, and reason about GUI actions in open-ended environments.
Recently, there has been a surge in the development of GUI-specific agents, which fine-tune MLLMs or build on their architecture to improve performance in structured GUI environments. Notable examples include:
CogAgent~\cite{hong2023cogagent}, built on the GLM-VLM~\cite{glm2024chatglm} framework, which achieves improvements in perception, action space coverage, and reasoning via staged training;
ShowUI~\cite{lin2024showui}, trained on top of Qwen2-VL~\cite{Qwen2-VL}, with enhancements for UI tokenization and cross-system navigation;
UI-TARS~\cite{qin2025ui}, which integrates Qwen2-VL~\cite{Qwen2-VL} with additional task planning and grounding modules, excelling in complex GUI manipulation tasks.
In particular, two fundamental capabilities have emerged as essential for effective GUI automation: element grounding, which refers to the ability to accurately locate relevant UI elements on the screen, and task completion, which involves determining the correct type and content of interaction at each step.

\subsection{GUI Benchmarks}

Benchmark datasets are fundamental to the development and evaluation of GUI agents. A key capability required for effective GUI interaction is accurate element grounding. For this purpose, ScreenSpot\cite{cheng2024seeclick} serves as a dedicated benchmark, offering both mobile and web-based examples to evaluate agents' spatial grounding capabilities. For task completion, Mind2Web\cite{deng2023mind2web} stands out as a pioneering benchmark built in real-world web environments rather than simplified simulations. WebVLN\cite{chen2024webvln} focuses on navigation and QA within shopping websites, while WebLINX\cite{lù2024weblinx} introduces multi-turn interactions in realistic web interfaces. Recently, some datasets have emerged that move toward dynamic GUI environment. OSWorld\cite{OSWorld} and WindowsAgentArena\cite{bonatti2024windows} are two benchmark platforms targeting GUI agents in Ubuntu and Windows operating systems, respectively. However, these dynamic testing are time-consuming and demand high setup Cost. In domain-specific research, WebWalkerQA\cite{wu2025webwalker} emphasizes web-based information retrieval and question answering. On the Android platform, PIXELHELP\cite{seq2act} and AITW (Android in the Wild)\cite{rawles2023androidwildlargescaledataset} offer static mobile task datasets, while META-GUI\cite{sun2022meta} focuses on multi-turn dialog interactions. GUI Odyssey\cite{lu2024gui} proposes cross-app task scenarios. Despite these advances, existing datasets are mostly constructed under ideal conditions, failing to address the irregularities and exceptions commonly found in real-world industrial scenarios. These abnormal situations are critical for evaluating agent \textbf{robustness}. WorldGUI\cite{zhao2025worldguidynamictestingcomprehensive} attempts to address this by exploring the impact of varying initial states, but many real-world anomalies remain unaddressed.
To fill this gap, we introduce \textbf{GUI-Robust}, a benchmark specifically designed to evaluate agent robustness under abnormal GUI scenarios.

\subsection{Data Collection Method for GUI Benchmark}

Most existing GUI datasets are collected with a manual pipeline ranging from task design to execution planning and manual demonstration. For instance, Mind2Web\cite{deng2023mind2web} outlines a process involving \textit{Website Selection}, \textit{Task Proposal}, \textit{Task Demonstration}, and \textit{Task Verification}, all of which heavily depend on human effort. Similarly, WebLINX\cite{lù2024weblinx} requires annotators to manually design tasks and engage in complex dialogues or task edits. To improve efficiency, some datasets leverage LLMs to accelerate task designs. Mind2Web\cite{deng2023mind2web} uses ChatGPT to generate seed tasks that inspire human annotators. GUI Odyssey\cite{lu2024gui} employs a template-based approach, where annotators design reusable templates and GPT-4 replaces entities (\eg app or item names) to generate multiple variants. Remarkably, these methods are still labor-extensive as it only accelerates task design procedure and still requires annetors for actions planning and execution, which is time-consuming.
A more autonomous approach is seen in WebWalkerQA\cite{wu2025webwalker}, where GPT-4o\cite{openai2024gpt4ocard} generates QA pairs based solely on webpage screenshots. While promising, this method is currently limited to QA-type tasks and does not generalize to broader GUI tasks.
To overcome these limitations, we propose a {reverse data collection paradigm} that collects user action sequences from natural interactions and then generate corresponding step and task descriptions for these actions with the assistance of MLLMs. Human annotators are only required for reviewing and revising step and task descriptions, which is highly efficient.  

\section{GUI-Robust Dataset}

\subsection{Overview}
\label{sec:dataset-composition}

GUI-Robust comprises 5,318 annotated tasks collected from 392 distinct websites and desktop applications. Each task (\ie data instance) consists of a task description and a sequence of user behaviors. Notably, GUI-Robust includes 200 tasks under abnormal scenarios, encompassing seven distinct types of anomalies commonly encountered in everyday GUI interactions. This enables rigorous evaluation of GUI agent robustness under real-world conditions. In addition to this unique feature, GUI-Robust offers a greater diversity and complexity of tasks, such as those involving information retrieval and cross-scenario operations. Furthermore, our dataset spans a wide range of 49 domains and covers nearly all commonly used platforms within the Chinese internet ecosystem. These characteristics ensure that GUI-Robust is closely aligned with real-world usage, thereby supporting comprehensive assessment of agent robustness, adaptability, and generalization. The dataset statistics are presented in Table ~\ref{tab:benchmark-comparison} and ~\ref{tab:dataset-composition}. 

To promote ease of use, we release a lightweight evaluation toolkit for GUI-Robust (see details in Appendix ~\ref{appendix:toolkit}), which enables users to quickly evaluate existing models or integrate their own with minimal effort.

\begin{table}[t]
    \centering
    \caption{Composition of the GUI-Robust Dataset}
    \label{tab:dataset-composition}
    \resizebox{\textwidth}{!}{
    \begin{tabular}{llll}
        \toprule
        \textbf{Category} & \textbf{Description} & \textbf{\#Tasks} & \textbf{Action Space} \\
        \midrule
        Standard Tasks               & routine tasks with single web/app       & 4414 & click, input \\
        Information Retrieval Tasks & tasks with information retrieval        & 503        & click, input, get\_info \\
        Cross-Scenario Tasks        & tasks across two webs/apps              & 201        & click, input, get\_info, open \\
        Abnormal Tasks              & tasks with abnormal scenarios           & 200  & click, input, get\_info, wait, human \\
        \bottomrule
    \end{tabular}
    }
\end{table}

\subsection{Metadata Definition}

\paragraph{Task Formulation.}
For a specific task, the execution of by a GUI agent can be represented as a sequence of actions paired with corresponding UI element coordinates at each step, as an example illustrated in Fig. ~\ref{fig:example}. Formally, given a task description $T$ and a screenshot $S_i$ at step $i$, the agent $\mathcal{G}$ generates an action $A_i$ and selects a UI element $E_i = (x_i, y_i)$ located at screen coordinates $(x, y)$. This step-wise interaction is defined as:
\[
R_i = (A_i, E_i) = \mathcal{G}(S_i, T)
\]
Upon task completion, the full execution result is represented as a sequence:
\[
R = \left\{(A_i, E_i)\,\middle|_ {i=1}^{n}, T \right\}
\]
where $n$ denotes the total number of steps required to complete the task. 

\paragraph{Dataset Structure.} GUI-Robust is composed of a collection of annotated tasks, each comprising a task description $T$  and a sequence of screenshots $S = \{S_1, S_2, \ldots, S_n\}$. The task description $T$ is provided in {natural language}, reflecting realistic user instructions (\eg \textit{"Search for the best-selling comic book on Amazon"}), while the sequence $S$ represents the screenshots captured at each interaction step. For each screenshot $S_i$, we also provide the corresponding ground-truth step description (\eg \textit{"Click on the 'Account \& Lists' text"}), the action performed (\eg \textit{"Click"}), and the element location. These annotations facilitate the evaluation of the agent's element grounding and task completion capabilities. See Appendix ~\ref{appendix:dataset-format} for details on the data format structure.



\paragraph{Element Type}
In this dataset, we categorize UI elements into three types: \textbf{icon}, \textbf{text}, and \textbf{box}. The icon type represents interactive UI elements that are abstract icons with no textual content. The text type refers to UI elements containing text, such as buttons or links. The box type represents input fields, where the task involves entering content into a text input box. This classification allows us to assess the localization capabilities of different models across various types of UI elements, providing insights into their performance in identifying and interacting with different element categories.

\paragraph{Action Space}
The action space in \textbf{GUI-Robust} comprises six distinct action types: \texttt{click}, \texttt{input}, \texttt{get\_info}, \texttt{open}, \texttt{wait}, and \texttt{human}. Among these, \texttt{click}, \texttt{input}, \texttt{get\_info} and \texttt{open} are four standard actions, corresponding to \textit{``click an element"}, \textit{``input the content"},  \textit{``retrieve information from the interface''}, \textit{``open a the website or application''}. In addition to these standard actions, GUI-Robust incorporates two specialized action types to address anomalous scenarios: \texttt{wait} denotes the \textit{``wait for response''} and \texttt{human} denotes the \textit{``require human intervention''}. the more details about these actions refer to Appendix ~\ref{appendix:dataset-format}.



\subsection{Abnormal Scenarios}
\label{sec:abnormal-scenario}
By collecting and analyzing 5,318 real user interaction sessions across a diverse range of websites and desktop applications, we identified seven types of commonly encountered anomalies (See Fig. \ref{fig:abn-sce}):

\begin{itemize} [leftmargin = *]
    \item \textbf{Action Failure:} The agent's previous action does not trigger the expected UI response (\eg a button click has no effect).
    \item \textbf{Login Page:} A login prompt appears unexpectedly, requiring authentication before proceeding.
    \item \textbf{Captcha Page:} The agent encounters a CAPTCHA challenge that it cannot autonomously solve.
    \item \textbf{Ad Pop-up:} Advertisements pop up and obscure key UI elements, disrupting the intended interaction flow.
    \item \textbf{Cookie Consent Pop-up:} A cookie consent dialog appears and must be dismissed before interacting with the main interface.
    \item \textbf{Page Loading Delay:} The page remains in a loading state for an extended time, preventing access to target elements.
    \item \textbf{Network Disconnection:} The interface fails to load due to temporary or complete loss of network connectivity.
\end{itemize}

These scenarios emerged organically from natural human activity on the internet and reflect common disruptions that hinder successful task completion. We introduce \texttt{wait} and \texttt{human} actions for agents to solve or propose the problem (See Table ~\ref{tab:abnormal-Scenarios}). This observation motivated us to explicitly incorporate these real-world failure cases into our dataset design, in order to enable more robust benchmarking and guide the development of agents that can operate reliably in practical environments.









\section{Data Collection Method}
\label{sec:data-collection-method}

\subsection{Existing Limitations}

Both step and task descriptions and corresponding action sequences are recorded to construct the dataset. However, this approach is labor-intensive and inefficient, particularly in the following two respects: 1) Task design typically depends on human heuristics, making the process cognitively demanding and time-consuming. Furthermore, the scope of manually designed tasks is often limited, and the curated task patterns frequently diverge from real-world user behaviors. 2) For recording action sequences, human annotators are required to have substantial familiarity with the target websites and desktop applications, which necessitates specialized training and often multiple iterations to produce accurate records. Even for annotators with considerable expertise, our experiments demonstrate that the average annotation time per task exceeds 15 minutes.



\subsection{New Data Collection Strategy:  RevAct }

To address these challenges, we propose a new data collection pipeline that reverses the convention
workflow and enables semi-automated dataset generation. The basic idea is to collect user action sequences from natural interactions, and then generate specific step and task descriptions for these actions with the assistance of MLLMs. This strategy could avoid the time-consuming steps of the task design and mannual demonstration, requiring Human annotators only involves to review and revise step and task descriptions generated by MLLMs. The pinepline is illustrated in Fig.~\ref{fig:upsidedown-pipeline}. Specifically, the data collection consists of the following three steps: 


\begin{figure}[t]  
    \centering  
    \includegraphics[width=\textwidth]{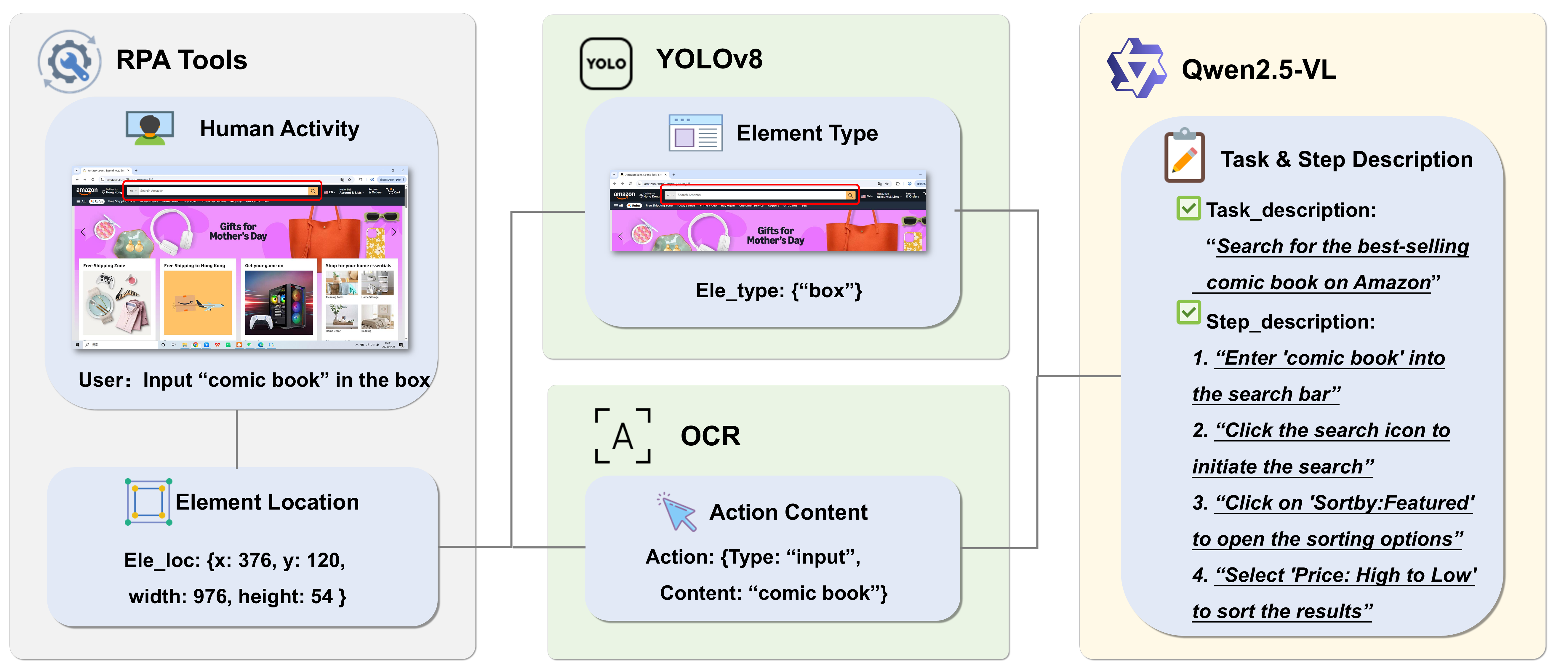}
    \caption{Overview of the RevAct pipeline for semi-automated data collection. Human interaction traces are captured by RPA tools, then processed by a YOLOv8 detector and OCR to extract element type, location, and action content. A multimodal LLM (Qwen2.5-VL) generates step-wise and task-level descriptions, enabling efficient and scalable construction of annotated GUI tasks.}  
    \label{fig:upsidedown-pipeline}  
     \vspace{-0.4cm}
\end{figure}

\begin{enumerate} [leftmargin = *]
    \item \textbf{Capture --- Screenshot and Element Coordinate Capture:} 
    We begin by collecting user action sequences derived from natural interactions. Specifically, we recruit volunteers and record their routine activities on familiar websites and desktop applications. During these sessions, a Robotic Process Automation tool (Indeed-Intelligence RPA\footnote{\url{https://www.ai-indeed.com/}}) automatically captures both screenshots and the coordinates of UI elements for each user action. In contrast to traditional approaches, which require annotators to perform pre-defined tasks demanding expertise and extensive planning, our data collection process is high efficient and does not necessitate specialized knowledge. This step can be easily and efficient completed for various volunteers without expertise, does not require expertise annotators. The economic cost associated with this procedure is negligible. Besides, this method ensures that the collected data more accurately reflects real-world user behavior. 
    
    

    \item \textbf{Interpretation --- Action Recognition:} The recorded screenshots and element coordinates are fed into a YOLOv8-based GUI element detection model, which identifies the type of UI element interacted with at each step. Subsequently, we establish a one-to-one mapping between element types and action types. For example:
    \begin{itemize}
        \item If the operated UI element is a clickable \texttt{icon} or \texttt{text}, the action type is mapped to \texttt{click}.
        \item If the UI element is an input \texttt{box}, the action type is mapped to text \texttt{input}. 
    \end{itemize}
   Unfortunately, we observed that YOLOv8 demonstrates limited accuracy in recognizing other types of actions. However, the number of other action types is relatively small, and we simply leave them for subsequent revision. Finally, we use an OCR tool to extract the content of the UI element, supplementing the action’s semantic information.

    \item \textbf{Summarize --- Step and Task Description Generation}: We feed the previously collected screenshots and the identified actions into Qwen2.5-VL~\cite{Qwen2.5-VL}. The model generates a step-by-step description based on this information, then summarizes all step descriptions to produce a task-level description, thereby completing the construction of a fully annotated data instance. In the prompt provided to the multimodal model, we explicitly instruct it to follow two key principles:
    (1) The task description should be a natural language summary of the overall task goal, rather than a simple concatenation of the step descriptions.
    (2) The description should emphasize the outcome of the final step, as this often captures the core objective of the task. At the same time, it should incorporate relevant details from intermediate steps to ensure that the resulting instruction is both semantically accurate and operationally complete.

     \item \textbf{Revise --- Refine the Step and Task Descriptions }: We employ human annotators to review and refine the step and task descriptions generated by MLLMs. Given that MLLMs exhibit relatively high accuracy, this process is efficient, as annotators are required to revise only a small subset of the descriptions.
     
\end{enumerate}

\subsection{Empirical Results}

To validate the effectiveness of the proposed method, we conducted a feasibility evaluation by having human annotators review and correct the task data generated by the RevAct method. The evaluation process focused on the following three aspects:

\begin{itemize} [leftmargin = *]
    \item \textbf{Accuracy of step descriptions}: Whether each step description accurately reflects the UI element and its content corresponding to the operation.
    \item \textbf{Consistency of task descriptions}: Whether the task description aligns with the actual intent and all steps.
    \item \textbf{Completeness of the task description}: Whether the task description contains sufficient information to reproduce the full interaction sequence—i.e., whether a human can follow it to reproduce the entire sequence of actions.
\end{itemize}

In the experiment, we test the descriptions generated by the RevAct method and invited 10 experienced annotators to independently correct and assess the quality. The results showed that, for standard tasks in Chinese, the step and task description accuracy was 78.63\% and 72.57\%. In comparison, for standard tasks in English, the step and task description accuracy reached 71.71\% and 86.67\%. These results indicate that the RevAct method can generate accurate operation and task text across different languages and task complexities.

In terms of data collection efficiency, RevAct significantly reduced the workload of annotators compared to traditional fully manual methods. Specifically, collecting and processing 100 task samples using traditional methods (See details in Appendix ~\ref{appendix:manual-data-collection}) typically requires 10 annotators and takes about 150 minutes, whereas the RevAct method, required only 10 annotators and 7.8 minutes to review and finalize the same number of collected samples. This represents an over 19$\times$ improvement in efficiency, highlighting the method's significant advantage in reducing human labor costs and accelerating data construction.


\section{Experiment}

\subsection{Experiment Setup}
\label{sec:expe-setup}

We evaluate 5 representative models on the \textbf{GUI-Robust} dataset, including 3 general-purpose multimodal large models with visual input and understanding capabilities—GPT4o~\cite{openai2024gpt4ocard}, Qwen2.5-VL~\cite{Qwen2.5-VL} and Gemini 2.5-Flash-Preview~\cite{geminiteam2024geminifamilyhighlycapable}—and two pretrained agents explicitly designed for GUI automation—UI-TARS~\cite{qin2025ui} and CogAgent~\cite{hong2023cogagent}. 

We design 2 evaluation types, comprising a total of 5 evaluation tasks:

\textbf{UI Element Grounding:} We randomly sample 1,028 single-step trajectories from the \textit{Standard Tasks} subset. Given a step description and its corresponding screenshot, models are evaluated on:
(a) \textbf{Action Accuracy (Action Acc.):} both action type and content must match the ground truth.
(b) \textbf{Coordinate Accuracy (Coord. Acc.):} predicted coordinates must fall within the bounding box of the target UI element.

\textbf{Task Completion:} We assigned 5 models to execute a set of tasks sampled from GUI-Robust: \textbf{302 Standard Tasks} (randomly selected from the 4,925-task subset) and \textbf{197 Abnormal Tasks}. During Task Completion evaluation, the model receives three inputs at each step:
(a) the task description, (b) the current page screenshot, and (c) a history of previously predicted actions and UI element coordinates.
Based on this context, the model is required to generate the next action and its corresponding element location, progressing until the entire task is completed. We evaluate: (a) Action Acc. (2) Coord. Acc. and (c) \textbf{Task Success Rate (SR):} the task is considered successful only if all actions and coordinates match the ground truth across the full trajectory.

\subsection{Results}

\begin{table}[t]
    \centering
    \caption{Performance on the UI Element Grounding. \textbf{Bold} indicates the best performance across all models, while \underline{underlined} values highlight the best among general-purpose multimodal models (MLLMs).}
    \label{tab:model-step-acc}
    \resizebox{0.9\textwidth}{!}{
        \begin{tabular}{l|c|cccc}
            \toprule
            \multirow{2}{*}{\textbf{Model}} & \multirow{2}{*}{\textbf{Action Acc.}} & \multicolumn{4}{c}{\textbf{Ele\_loc Acc.}} \\
            \cmidrule(lr){3-6}
            & & \textbf{Icon\_ele} & \textbf{Text\_ele} & \textbf{Box\_ele} & \textbf{All} \\
            \midrule
            Qwen2.5-vl-7B & 82.59 & 4.63 & \underline{5.07} & 4.13 & 4.77 \\
            Qwen2.5-vl-72B & 89.20 & \underline{5.56} & 3.21 & 1.83 & 3.40 \\
            Gemeini2.5-Flash-Preview & 77.24 & 1.85 & 3.72 & \underline{22.02} & \underline{7.20} \\
            GPT-4o-2024-11-20 & \underline{\textbf{90.56}} & 1.39 & 3.04 & 5.50 & 3.21 \\
            \midrule
            UI-TARS-1.5-7B & 51.17 & 76.85 & \textbf{84.63} & 62.39 & \textbf{78.31} \\
            CogAgent-9B & 42.40 & \textbf{78.65} & 67.67 & \textbf{89.50} & 77.12 \\
            \bottomrule
        \end{tabular}
    }
     \vspace{-0.4cm}
\end{table}

\paragraph{UI Element Grounding.} 
Table~\ref{tab:model-step-acc} presents the results of various models on the UI Element Grounding task. We draw several key observations: 

\begin{enumerate} [leftmargin = *]
    \item UI-TARS and CogAgent significantly outperform general-purpose MLLMs on grounding accuracy. UI-TARS achieves the highest overall localization accuracy (78.31\%) across all element types, followed closely by CogAgent (77.12\%). This demonstrates the advantage of GUI-specific pretrained models in accurately grounding UI elements, especially for visually complex or fine-grained components. All MLLMs underperform significantly in overall localization ($\leq$ 7.20\%), highlighting a persistent gap in visual grounding capabilities when applied to GUI domains; 

    \item General-purpose MLLMs like GPT-4o and Qwen2.5 show strong action recognition but weak spatial grounding. GPT-4o-2024 achieves the best action accuracy at 90.56\%, and Qwen2.5-vl-72B follows with 89.20\%. However, both models struggle to localize elements precisely—e.g., GPT-4o achieves only 3.21\% overall localization accuracy, suggesting that while it understands what to do, it often fails to locate where to do it; 

    \item  Different models specialize in different UI element types. CogAgent excels at localizing input boxes (89.50\%) and icon elements (78.65\%). UI-TARS performs best on text-based elements (84.63\%). Interestingly, Gemini2.5-Flash achieves the highest accuracy on box elements (22.02\%) among the MLLMs, despite lower overall performance.
\end{enumerate}


\paragraph{Task Completion.} 
Table~\ref{tab:normal_abnormal} presents a comparative analysis of model performance on full-task execution under both normal and abnormal conditions. Several key observations emerge:

\begin{enumerate} [leftmargin = *]

    \item Under normal conditions, GUI-specific agents outperform general-purpose models. 
    UI-TARS and CogAgent demonstrate significantly higher task success rates (21.61\% and 14.38\%, respectively) and element localization accuracy (62.33\% and 54.32\%) than LLM-based models such as GPT-4o or Qwen2.5-VL. This suggests that specialized training on GUI interactions yields clear benefits in structured, expected environments.

    \item In abnormal settings, general-purpose LLMs show stronger anomaly awareness, but poor spatial grounding.
    Despite lower task success rates, models like GPT-4o and Qwen2.5-VL maintain higher action accuracy (55.07\% and 48.54\%, respectively) compared to GUI-specific agents. This indicates that LLMs are more sensitive to abnormal cues (e.g., detecting login pop-ups or page errors), but fail to locate or interact with the appropriate UI elements—limiting their ability to recover or proceed correctly.

\end{enumerate}

\begin{table}[t]
    \centering
    \caption{Performance on Task Compeletion evaluation. \textbf{Bold} indicates the best performance across all models, while \underline{underlined} values highlight the best among general-purpose multimodal models (MLLMs).}
    \label{tab:normal_abnormal}
    \resizebox{\textwidth}{!}{
    \begin{tabular}{l|ccc|ccc}
        \toprule
        \multirow{2}{*}{\textbf{Model}} &
        \multicolumn{3}{|c}{\textbf{Normal}} &
        \multicolumn{3}{|c}{\textbf{Abnormal}} \\
        \cmidrule(lr){2-4} \cmidrule(lr){5-7}
        & \textbf{Action Acc.} & \textbf{Ele\_loc Acc.} & \textbf{Task SR} &
        \textbf{Action Acc.} & \textbf{Ele\_loc Acc.} & \textbf{Task SR} \\
        \midrule
        Qwen2.5-VL-72B & \underline{\textbf{56.99}} & 3.03 & {\underline{1.32}} & 48.54 & 0.98 & 0 \\
        Gemini2.5-flash-preview & 50.63 & \underline{6.78} & 0 & 45.15 & \underline{5.13} & 0 \\
        GPT-4o-2024-11-20 & 56.26 & 1.88 & 0.33 & \underline{\textbf{55.07}} & 1.95 & 0 \\
        \midrule
        UI-TARS-1.5-7B & 35.36 & \textbf{62.33} & \textbf{21.61} & 28.88 & \textbf{61.77} & \textbf{4.73} \\
        CogAgent-9B & 33.12 & 54.32 & 14.38 & 23.48 & 44.35 & 2.14 \\
        \bottomrule
    \end{tabular}
    }
     \vspace{-0.4cm}
\end{table}

GUI-specific agents struggle with unanticipated anomalies.
These models often lack mechanisms to recognize or adapt to unexpected interface changes. For example, during a task on the “58.com” website, UI-TARS encounters a sudden login page. Instead of detecting and reporting the abnormal transition, it predicts the action "click 58.com", which will cause an infinite redirection loop. This failure illustrates the rigidity of GUI-specific models when exposed to disruptions outside their training distribution.

Incorrect behavior under anomalies can lead to critical consequences.
In real-world applications, failure to handle abnormal states may result in hazardous operations such as accidental deletions, repeated entry of sensitive credentials, or navigation to malicious or unintended pages. Such behaviors not only reduce task reliability but also raise serious safety and data integrity concerns. These results highlight the need for robust anomaly-handling capabilities in GUI agents—not only to succeed in idealized environments, but also to detect, report, and gracefully recover from irregularities.

\section{Discussion}
\label{sec:discuss}

\paragraph{Limitations.} 
\label{sec:discuss-limit}
While GUI-Robust offers a substantial step toward evaluating agent robustness in realistic GUI environments, several limitations remain. The abnormal scenarios in our dataset—though diverse—are still limited to 7 predefined categories. Real-world applications may involve more complex or compound failure modes (e.g., cascading authentication prompts, adaptive overlays, operations with insufficient permissions), which are not currently covered. Second, GUI-Robust adopts a static dataset and evaluation paradigm, where each task is annotated with a single canonical execution trajectory. 

\paragraph{Future Work.}
Future extensions of GUI-Robust aim to address these limitations along two directions. First, we plan to enrich the spectrum of abnormal scenarios by including more diverse failure types—such as dynamic Captcha variants, pop-ups requiring multi-step dismissal, and access control violations. Second, we envision evolving GUI-Robust into a dynamic evaluation platform, where agents interact with simulated environments rather than static trajectories. In such settings, evaluation criteria can move beyond exact action or coordinate matching to focus on goal-oriented success, state transition correctness, and error recovery ability. This shift will better reflect real-world deployment conditions and enable the development of agents that are not only accurate but also flexible, adaptive, and robust.

\section{Conclusion}

In this work, we present GUI-Robust, a comprehensive benchmark designed to evaluate the robustness of GUI agents under real-world anomalies, which incorporates 7 types of common abnormal scenarios. We further introduce RevAct, an efficient upside-down data collection pipeline that leverages multimodal models to generate rich, naturalistic task annotations from user behavior traces. Through extensive experiments we demonstrate that existing GUI-specific agents struggle in failure-prone environments. These findings highlight the pressing need for robustness-focused datasets and evaluation protocols. We hope GUI-Robust will serve as a valuable resource for advancing the development of reliable, adaptable, and safe GUI agents in real-world applications.

\bibliographystyle{plainnat}
\bibliography{references}


\appendix

\section{Dataset Format}
\label{appendix:dataset-format}

Each datapoint is stored as a single JSON file accompanied by a set of PNG screenshots, one for each interaction step. The JSON file contains the following fields:

\begin{itemize} [leftmargin = *]
    \item \texttt{task\_description:} A natural language instruction describing the overall task goal the agent should complete.
    \item \textbf{A set of Steps:} Each task consists of multiple interaction steps, each containing the following fields:
        \begin{itemize}
            \item \texttt{step\_description:} A natural language description of the expected action at this step, grounded in the corresponding screenshot.
            \item \texttt{img\_path:} The file name to the screenshot image (PNG format) representing the current GUI state.
            \item \texttt{ele\_loc/x:} The x-coordinate of the top-left corner of the target UI element.
            \item \texttt{ele\_loc/y:} The y-coordinate of the top-left corner of the target UI element.
            \item \texttt{ele\_loc/width:} The width of the bounding box surrounding the target UI element.
            \item \texttt{ele\_loc/height:} The height of the bounding box surrounding the target UI element.
            \item \texttt{ele\_type:} The type of the UI element, categorized as \texttt{icon}, \texttt{text}, or \texttt{box}.
            \item \texttt{action/type:} The type of action to be performed, selected from \texttt{click}, \texttt{input}, \texttt{get\_info}, \texttt{open}, \texttt{wait}, or \texttt{human}.
            \item \texttt{action/content:} The semantic content of the action, such as the text to input or the label of a button to click.
            \item \texttt{action\_human (optional):} For abnormal scenarios, this field provides example content for the \texttt{human} action type, describing the issue that the agent should report to a human operator.
        \end{itemize}
\end{itemize}

Table~\ref{tab:action-space} defines the complete action space used in GUI-Robust. Each step in a task is annotated with an action composed of two components: a type, which specifies the kind of user interaction (e.g., click, input), and a content, which provides semantic detail about the interaction (e.g., the text to input or the target element to click):

\begin{itemize} [leftmargin = *]
    \item \texttt{click}: Represents a standard mouse click operation. The \texttt{content} field describes the semantic label or textual content of the clicked element (if available), such as a clickable text labeled “Search” or “Next”.

    \item \texttt{input}: Denotes a text input operation into a form field or search bar. The \texttt{content} provides the actual text string to be entered, such as “comic book” in a search box.

    \item \texttt{get\_info}: Indicates an information retrieval action. The \texttt{content} describes the specific piece of information to be extracted from the GUI, such as a price, a user name, or a product description.

    \item \texttt{wait}: A passive action used when the GUI is temporarily unresponsive (e.g., during page loading). This type has no content field, as the action itself signals a need to pause before proceeding.

    \item \texttt{open}: Represents an explicit instruction to launch a new web page or desktop application. The \texttt{content} field specifies the name or identifier of the target app or site to be opened.

    \item \texttt{human}: Used when the agent encounters an abnormal scenario it cannot resolve autonomously, such as login screens or Captcha challenges. The \texttt{content} field provides a human-readable description of the issue, serving as a report for fallback intervention.
\end{itemize}

\begin{table}[ht]
    \centering
    \caption{Action space(type + content) and their semantic meanings}
    \label{tab:action-space}
    \resizebox{\textwidth}{!}{
    \begin{tabular}{lll}
        \toprule
        \textbf{Type} & \textbf{Content} & \textbf{Description} \\
        \midrule
        \texttt{click} & The content of the clicked element (if any) & A click action. \\
        \texttt{input} & The text to be entered. & A text input action. \\
        \texttt{get\_info} & The information be retrieved. & An operation to retrieve information from the interface; \\
        \texttt{wait} & None & Signifies a wait action \\
        \texttt{human} & A description of the problem encountered by agent. & An action to report the abnormal scenario. \\
        \texttt{open} & The name of the web/app to open. & Action to open a web/app \\
        \bottomrule
    \end{tabular}
    }
\end{table}

\section{Data Example}

\begin{figure}[ht]  
    \centering  
    \includegraphics[width=\textwidth]{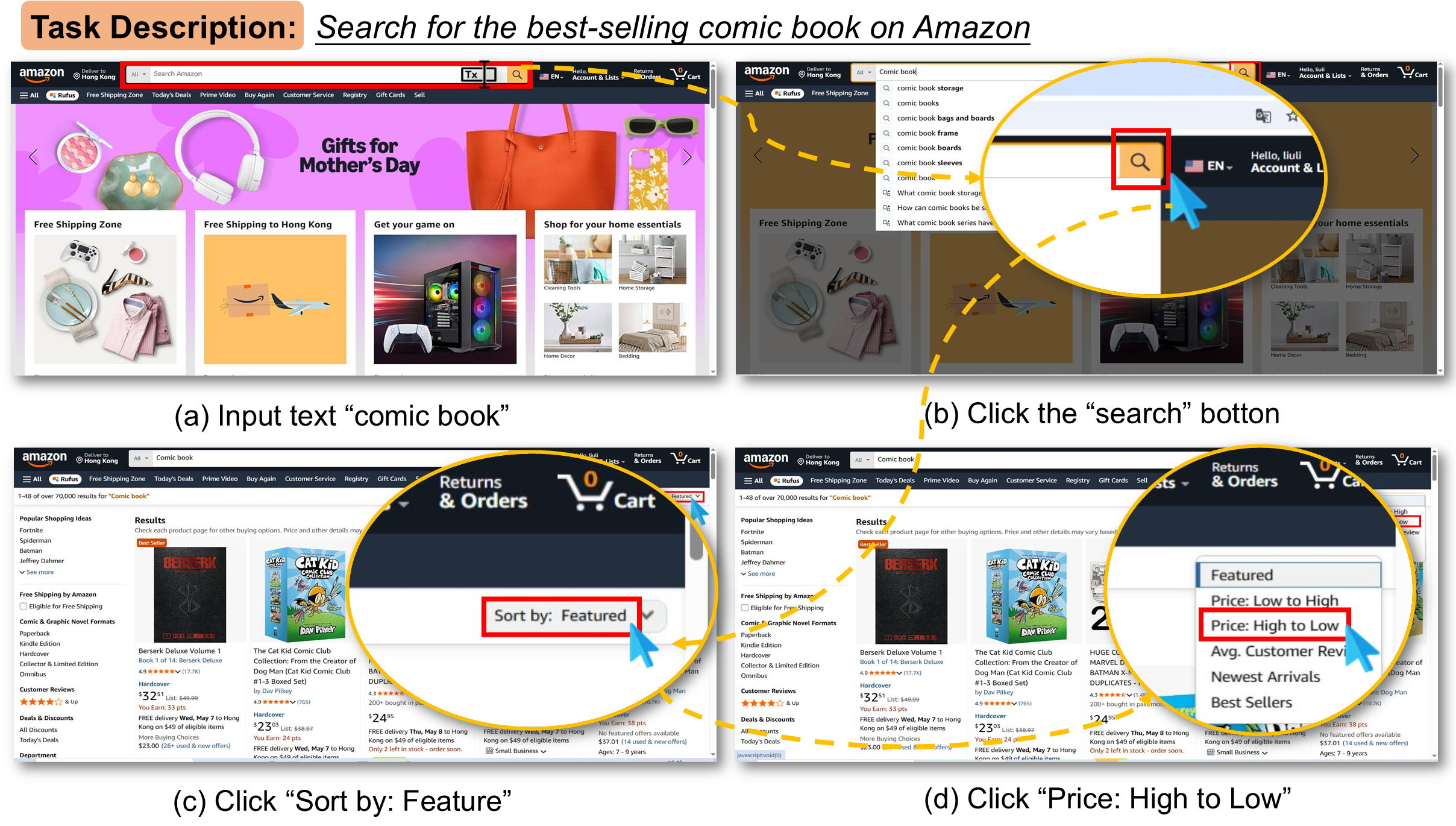}
    \caption{An example task from GUI-Robust: Search for the best-selling comic book on Amazon. The task is decomposed into four steps: (a) enter the query “comic book” in the search bar, (b) click the search button, (c) open the sorting dropdown, and (d) select the “Price: High to Low” option. Each step corresponds to a user interaction grounded in the GUI screenshot.}  
    \label{fig:example}  
\end{figure}

See an example in Fig. ~\ref{fig:example}

\section{Abnormal Scenarios Example}

In this appendix, we provide illustrative examples and formal definitions of abnormal GUI scenarios included in the GUI-Robust dataset. These scenarios reflect common disruptions that frequently occur in real-world GUI usage but are largely overlooked in prior benchmarks.

\begin{figure}[ht]  
    \centering  
    \includegraphics[width=\textwidth]{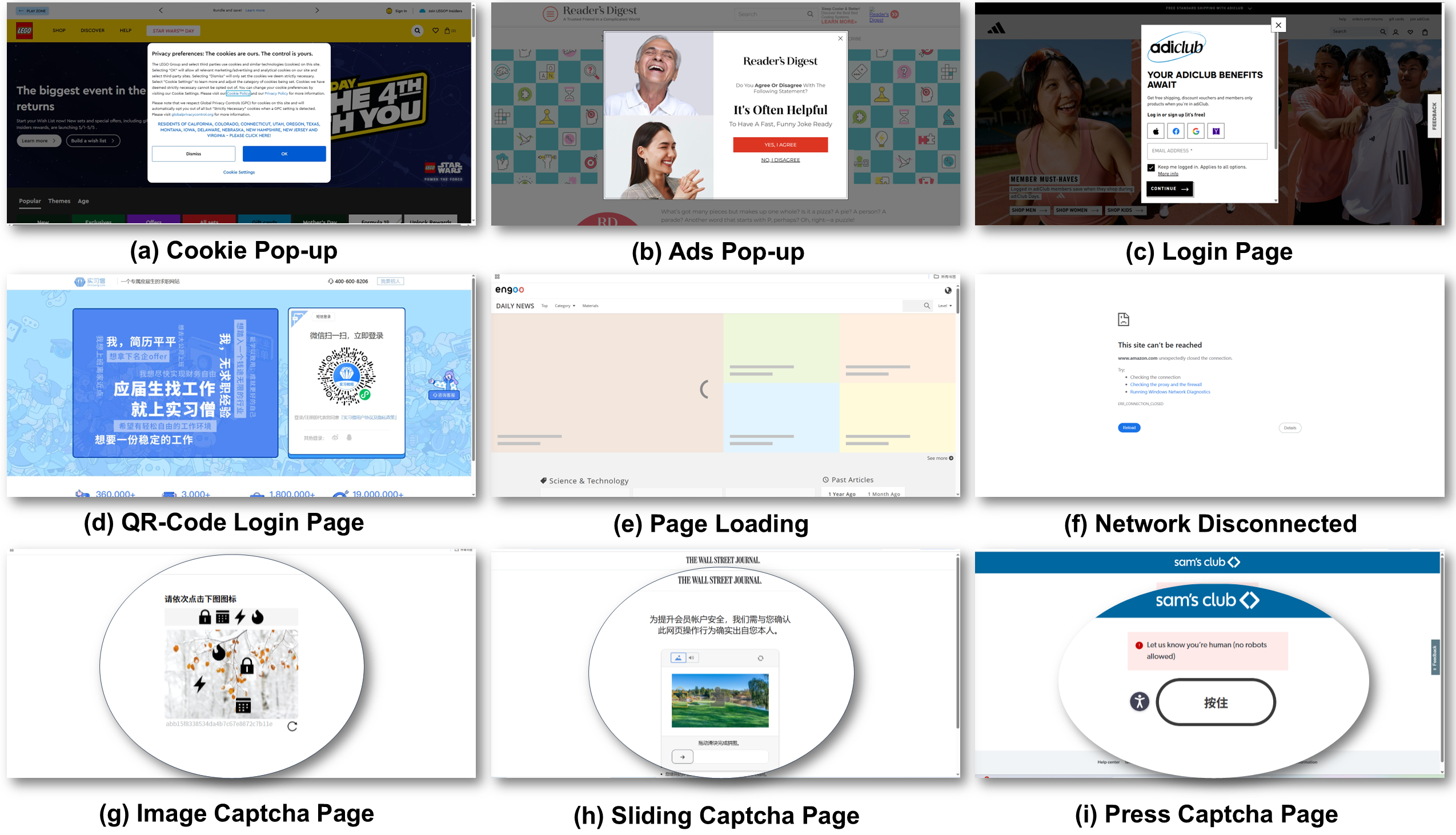}
    \caption{Examples of abnormal GUI scenarios in the GUI-Robust dataset. (a) cookie pop-up, (b) advertisement pop-up, (c) login page, (d) QR-code login prompt, (e) page loading state, (f) network disconnection, (g) image-based Captcha, (h) sliding Captcha, and (i) press-to-verify Captcha. These scenarios require agents to detect and appropriately handle interruptions in order to maintain task robustness.}  
    \label{fig:abn-sce}  
\end{figure}

Fig.~\ref{fig:abn-sce} presents visual examples of nine representative abnormal conditions. These include transient interface elements such as cookie consent dialogs (a) and advertisement pop-ups (b), as well as system-level disruptions such as login prompts (c, d), Captcha challenges (g–i), page loading delays (e), and network disconnections (f). Each of these anomalies can significantly interrupt the agent's execution flow and requires appropriate detection and response strategies.

Table~\ref{tab:abnormal-Scenarios} summarizes the expected agent behaviors in response to each abnormal scenario. For instance, when encountering a login or Captcha page, the agent should generate a \texttt{human} action to report that manual intervention is required. In contrast, if a pop-up can be resolved autonomously (e.g., cookie acceptance or ad dismissal), the agent should attempt to issue a valid \texttt{click} action on the relevant UI element. In some cases, such as delayed page loading, a \texttt{wait} action is necessary before retrying the intended operation.

\begin{table}[t]
    \centering
    \caption{Abnormal Scenarios and Expected Actions}
    \label{tab:abnormal-Scenarios}
    \resizebox{\textwidth}{!}{
    \begin{tabular}{ll}
        \toprule
        \textbf{Abnormal Scenario} & \textbf{Expected Action} \\
        \midrule
        Action Failure & 1. human: Detected failure of the previous action;\\ & 2. click/input (repeat previous action) \\
        Login Page & human: Detected the login page, which needs to be completed manually \\
        Captcha Page & human: Detected the captcha page, which needs to be completed manually \\
        Ad Pop-up & 1. human: Detected an ad pop-up blocking the process;\\ & 2. click (close the pop-up) \\
        Cookie Pop-up & 1. human: Detected a cookie pop-up blocking the process;\\ & 2. click (accept cookies) \\
        Page Loading & 1. human: Detected that the page is loading;\\ & 2. wait \\
        Network Disconnection & human: Detected that the network is currently disconnected \\
        \bottomrule
    \end{tabular}
    }
\end{table}

\section{Toolkit Integration Guide}
\label{appendix:toolkit}

This appendix provides details on how to integrate custom models into the evaluation pipeline.

\paragraph{Model Interface Specification} 
Each model should implement the following interface:

\begin{verbatim}
class YourModel:
    def __init__()
    
    def pred_step_loc(self, step_description: str, screenshot_base64: str) \
        -> dict:
        # Return a prediction for the current step
        return {
            "ele_loc": {"x": 100, "y": 200},
            "ele_type": "text",
            "action": {"type": "click", "content": "Search"}
        }

    def pred_task_full(self, task_description: str, \
        screenshot_list_base64: List[str]) -> List[dict]:
        # Return a list of predictions for the full task
        return [ ... ]  # One entry per step
\end{verbatim}

These two methods respectively handle single-step prediction and full-task multi-step prediction. The evaluation pipeline will automatically invoke these methods, compare predictions against ground truth, and report metrics such as action accuracy, coordinate accuracy, and task success rate.

\paragraph{Prediction Output Format} Each prediction should contain:
\begin{itemize}
    \item Element coordinates (\texttt{x, y})
    \item Element type (\texttt{icon, text, box, none})
    \item Action type and content (\texttt{click, input, get\_info, open, wait, human})
\end{itemize}
For full-task predictions, return a list of such dictionaries (one per step).

\paragraph{Evaluation Script} Run the toolkit via:
\begin{verbatim}
python evaluation.py \
  --model_name YourModel \
  --eval_type step|task \
  --task_type normal|abnormal \
  --data_path path_to_data_folder
\end{verbatim}

The evaluation script supports two modes:

\texttt{--eval\_type step}: Evaluates per-step grounding accuracy (Action acc. and Coord. acc.)

\texttt{--eval\_type task}: Evaluates full task execution (Action acc., Coord. acc., and Task Success)

\paragraph{Prompt Templates}
The evaluation toolkit internally uses task-specific prompt templates for multimodal models. For example:

\begin{lstlisting}

You are a GUI agent. You are given a step with screenshot. 
You need to perform the next action to complete the task.

## Output Format
```
{
    "ele_loc": "(x, y)",
    "ele_type": "",
    "action": {
        "type": "",
        "content": ""
    }
}
```

## action Space

| type | content | Description                |
| -------- | ------------------------------------------------------------ |  ------------------------------------------------------------ |
| click | The content of the element clicked (if any), for example, when clicking on a text element, content is the text content clicked | click |
| input | Input text | Input text         |
| get_info | Obtained information | Information acquisition operation, some operations involve obtaining information in the interface |


## ele_loc Space

| ele_type | Description |
| -------- | ------------------------------------------------------------ |
| icon | Abstract icon class element, that is, an icon without text |
| text | Text class element, that is, an icon containing text |
| box | Elements such as input boxes |
| none | has no element type. The action get_info does not correspond to any UI element, so the type is none |

## Note
- Use Chinese if the 'content' is Chinese. And use English if the 'content' is English.

\end{lstlisting}

Our codebase is publicly available at:
\begin{center}
  \url{https://github.com/chessbean1/GUI-Robust}
\end{center}

\section{Details of Manual Data Collection Pipeline}
\label{appendix:manual-data-collection}

To provide a meaningful baseline for comparison, we also implemented a traditional fully manual data collection pipeline. This five-step approach mirrors common practices used in existing GUI datasets and serves as a contrast to our proposed semi-automated RevAct pipeline. The workflow is as follows:

\paragraph{Step 1: Task Design.} Annotators first select a specific application or website and design a meaningful task based on real-world usage scenarios (e.g., "check the latest train schedule", "send a message via the chat app").

\paragraph{Step 2: Step Decomposition.} Each task is decomposed into atomic steps, corresponding to the minimal sequence of user interactions required to complete the task. Each step should reflect a single action on a clearly identifiable UI element.

\paragraph{Step 3: Screenshot Capture.} For every step, annotators manually take a screenshot of the GUI interface immediately before the interaction occurs. These screenshots serve as the visual grounding context for the agent.

\paragraph{Step 4: Annotation.} Annotators label each step with the following fields: a natural language step description, element location (bounding box coordinates), element type, and action type/content. For abnormal tasks, the corresponding \texttt{human} action content is also specified.

\paragraph{Step 5: Review and Correction.} A second annotator independently checks each task for correctness and consistency. Common issues include incorrect coordinates, ambiguous step descriptions, or wrong actions. Such issues are manually flagged and revised to ensure annotation quality.

In practice, we observed several challenges associated with the traditional manual data collection process, which significantly limit its scalability and consistency. These include:

\begin{itemize} [leftmargin = *]

    \item \textbf{High Learning Overhead for New Applications:} Annotators often face steep learning curves when unfamiliar with the target software. Understanding the interface structure, navigation flow, and interaction logic may require extensive exploration, increasing the overall collection time and cognitive burden.
    
    \item \textbf{Annotation Complexity and Quality Sensitivity:} In addition to executing tasks, annotators must carefully label each step with precise formatting. They must also pay attention to dynamic interface states, such as pop-up dialogs, cookie banners, or login modals, which may partially or fully obscure the target UI elements.
    
    \item \textbf{Human Variability and Subjective Interpretation:} Different annotators may interpret the same task or GUI state differently, leading to inconsistencies in task decomposition, step description, or element labeling. Without strict guidelines and training, this can introduce annotation noise and reduce benchmark reliability.
    
    \item \textbf{Error Propagation and Correction Overhead:} Errors made early in the process—such as a misinterpreted task goal or incorrect screenshot timing—can propagate through all subsequent steps. Detecting and fixing these mistakes requires full reprocessing of the trajectory, making error correction costly and inefficient.

\end{itemize}

These limitations highlight the practical difficulties of large-scale, high-quality GUI annotation when relying solely on manual pipelines. They further motivate the need for semi-automated approaches such as RevAct that can reduce human burden while preserving annotation richness and realism.


\newpage
\section*{NeurIPS Paper Checklist}

\begin{enumerate}

\item {\bf Claims}
    \item[] Question: Do the main claims made in the abstract and introduction accurately reflect the paper's contributions and scope?
    \item[] Answer: \answerYes{} 
    \item[] Justification: The contribution of this paper is summarized in the end of Introduction.

\item {\bf Limitations}
    \item[] Question: Does the paper discuss the limitations of the work performed by the authors?
    \item[] Answer: \answerYes{} 
    \item[] Justification: We discuss the limitation of this paper in Section \ref{sec:discuss}

\item {\bf Theory assumptions and proofs}
    \item[] Question: For each theoretical result, does the paper provide the full set of assumptions and a complete (and correct) proof?
    \item[] Answer: \answerNA{} 
    \item[] Justification: Our work focuses on a new dataset construction, therefore does not include theoretical results. 

    \item {\bf Experimental result reproducibility}
    \item[] Question: Does the paper fully disclose all the information needed to reproduce the main experimental results of the paper to the extent that it affects the main claims and/or conclusions of the paper (regardless of whether the code and data are provided or not)?
    \item[] Answer: \answerYes{} 
    \item[] Justification: We have released the evaluation scripts on GitHub, and provide detailed usage instructions in Appendix~\ref{appendix:toolkit}.

\item {\bf Open access to data and code}
    \item[] Question: Does the paper provide open access to the data and code, with sufficient instructions to faithfully reproduce the main experimental results, as described in supplemental material?
    \item[] Answer: \answerYes{} 
    \item[] Justification: We have uploaded the code at \url{https://github.com/chessbean1/GUI-Robust}

\item {\bf Experimental setting/details}
    \item[] Question: Does the paper specify all the training and test details (e.g., data splits, hyperparameters, how they were chosen, type of optimizer, etc.) necessary to understand the results?
    \item[] Answer: \answerYes{} 
    \item[] Justification: The detailed experimental settings are introduced in section ~\ref{sec:expe-setup} and Appendix ~\ref{appendix:toolkit}.

\item {\bf Experiment statistical significance}
    \item[] Question: Does the paper report error bars suitably and correctly defined or other appropriate information about the statistical significance of the experiments?
    \item[] Answer: \answerNo{} 
    \item[] Justification: We report deterministic performance metrics for all models based on fixed test sets without repeated trials or variance analysis; thus, no statistical significance or error bars are provided.

\item {\bf Experiments compute resources}
    \item[] Question: For each experiment, does the paper provide sufficient information on the computer resources (type of compute workers, memory, time of execution) needed to reproduce the experiments?
    \item[] Answer: \answerYes{} 
    \item[] Justification: All experiments were conducted on A100 GPUs (40GB), using either 1 or 2 GPUs depending on model size (e.g., UI-TARS-7B vs. Qwen2.5-VL-72B). Each evaluation run on our benchmark typically took 20–60 minutes per model. We estimate the total compute budget for all reported experiments to be approximately 150 GPU-hours.
    
\item {\bf Code of ethics}
    \item[] Question: Does the research conducted in the paper conform, in every respect, with the NeurIPS Code of Ethics \url{https://neurips.cc/public/EthicsGuidelines}?
    \item[] Answer: \answerYes{} 
    \item[] Justification: Our research complies with the NeurIPS Code of Ethics. All data were collected from publicly accessible platforms with no personally identifiable information.

\item {\bf Broader impacts}
    \item[] Question: Does the paper discuss both potential positive societal impacts and negative societal impacts of the work performed?
    \item[] Answer: \answerYes{} 
    \item[] Justification: Our dataset supports the development of more robust and generalizable GUI agents, which could reduce human workload in software operation, accessibility, and automation. 
    
\item {\bf Safeguards}
    \item[] Question: Does the paper describe safeguards that have been put in place for responsible release of data or models that have a high risk for misuse (e.g., pretrained language models, image generators, or scraped datasets)?
    \item[] Answer: \answerNA{} 
    \item[] Justification: Our work does not involve releasing models or data with high risk of misuse. The dataset contains only GUI interaction traces from publicly accessible applications and does not include sensitive content or private user information.

\item {\bf Licenses for existing assets}
    \item[] Question: Are the creators or original owners of assets (e.g., code, data, models), used in the paper, properly credited and are the license and terms of use explicitly mentioned and properly respected?
    \item[] Answer: \answerYes{} 
    \item[] Justification: All external assets used in this work—such as pretrained models (e.g., Qwen2.5-VL) and benchmarks (e.g., Mind2Web, WebLINX)—are properly cited with references to their original papers. We adhere to their respective open-source licenses (e.g., Apache 2.0, CC-BY 4.0) where applicable, and no proprietary or restricted resources were used.

\item {\bf New assets}
    \item[] Question: Are new assets introduced in the paper well documented and is the documentation provided alongside the assets?
    \item[] Answer: \answerYes{} 
    \item[] Justification: We release a new dataset (GUI-Robust) and evaluation scripts, both of which are accompanied by detailed documentation.

\item {\bf Crowdsourcing and research with human subjects}
    \item[] Question: For crowdsourcing experiments and research with human subjects, does the paper include the full text of instructions given to participants and screenshots, if applicable, as well as details about compensation (if any)? 
    \item[] Answer: \answerNo{} 
    \item[] Justification: Our data collection involved in-house annotators rather than public crowdsourcing platforms. While we provided task guidelines and interface demonstrations, we did not include the full instruction text or compensation details in the paper or supplementary material.

\item {\bf Institutional review board (IRB) approvals or equivalent for research with human subjects}
    \item[] Question: Does the paper describe potential risks incurred by study participants, whether such risks were disclosed to the subjects, and whether Institutional Review Board (IRB) approvals (or an equivalent approval/review based on the requirements of your country or institution) were obtained?
    \item[] Answer: \answerYes{}{} 
    \item[] Justification: Our study does not involve research with human subjects as defined by IRB standards. All data were collected via non-identifiable interaction traces from internal annotators performing routine tasks on public applications.

\item {\bf Declaration of LLM usage}
    \item[] Question: Does the paper describe the usage of LLMs if it is an important, original, or non-standard component of the core methods in this research? Note that if the LLM is used only for writing, editing, or formatting purposes and does not impact the core methodology, scientific rigorousness, or originality of the research, declaration is not required.
    \item[] Answer: \answerYes{} 
    \item[] Justification: Our data collection pipeline (RevAct) leverages the Qwen2.5-VL multimodal large language model to generate step-wise and task-level natural language descriptions based on user interaction traces. This usage is integral to the core methodology and is described in detail in Section~\ref{sec:data-collection-method}. 

\end{enumerate}

\end{document}